\documentclass[letterpaper, 10 pt, conference]{ieeeconf}

\IEEEoverridecommandlockouts      

\usepackage{amsmath} 
\usepackage{graphicx}
\usepackage{booktabs}
\usepackage{amssymb}
\usepackage{siunitx}
\usepackage[english]{babel}
\usepackage[noadjust]{cite}
\usepackage{units}
\usepackage{epstopdf}
\usepackage[font=small]{caption}
\usepackage{subcaption}
\usepackage{comment}
\usepackage{amsfonts}
\usepackage{bm}
\usepackage{algorithm}
\usepackage{balance}

\usepackage{mathtools}
\usepackage{centernot}

\usepackage{amsthm}
\usepackage{algorithmic}

\usepackage{color}

\usepackage{xcolor}
\newcommand{\remove}[1]{}

\theoremstyle{plain}

\title{\LARGE \bf
Active Probing and Influencing Human Behaviors Via Autonomous Agents
}

\author{Shuangge Wang$^1$, Yiwei Lyu$^2$, John M. Dolan$^3$
\thanks{$^1$Shuangge Wang is with the Ming Hsieh Department of Electrical and Computer Engineering, University of Southern California, Los Angeles, CA, 90089 USA. Email: {\tt \small larrywan@usc.edu}}
\thanks{$^{2}$Yiwei Lyu is with the Department of Electrical and Computer Engineering, Carnegie Mellon University, Pittsburgh, PA 15213 USA. Email: {\tt \small yiweilyu@andrew.cmu.edu}}
\thanks{$^{3}$John M. Dolan is with the Robotics Institute, Carnegie Mellon University, Pittsburgh, PA 15213 USA. Email: {\tt \small jdolan@andrew.cmu.edu}}
}

\begin{document}
\maketitle
\thispagestyle{empty}
\pagestyle{empty}

\begin{abstract}
Autonomous agents (robots) face tremendous challenges while interacting with heterogeneous human agents in close proximity. One of these challenges is that the autonomous agent does not have an accurate model tailored to the specific human that the autonomous agent is interacting with, which could sometimes result in inefficient human-robot interaction and suboptimal system dynamics. Developing an online method to enable the autonomous agent to learn information about the human model is therefore an ongoing research goal. Existing approaches position the robot as a passive learner in the environment to observe the physical states and the associated human response. This passive design, however, only allows the robot to obtain information that the human chooses to exhibit, which sometimes doesn’t capture the human’s full intention. In this work, we present an online optimization-based probing procedure for the autonomous agent to clarify its belief about the human model in an active manner. By optimizing an information radius, the autonomous agent chooses the action that most challenges its current conviction. This procedure allows the autonomous agent to actively probe the human agents to reveal information that’s previously unavailable to the autonomous agent. With this gathered information, the autonomous agent can interactively influence the human agent for some designated objectives. Our main contributions include a coherent theoretical framework that unifies the probing and influence procedures and two case studies in autonomous driving that show how active probing can help to create better participant experience during influence, like higher efficiency or less perturbations.

\end{abstract}

\section{Introduction}
\begin{figure}
\begin{tabular}{ c c}
\includegraphics[width=10pt, angle = -90]{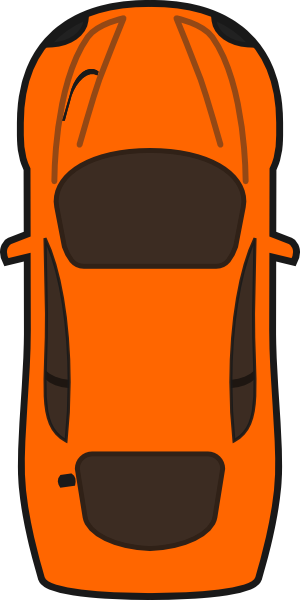} & \includegraphics[width=10pt, angle = -90]{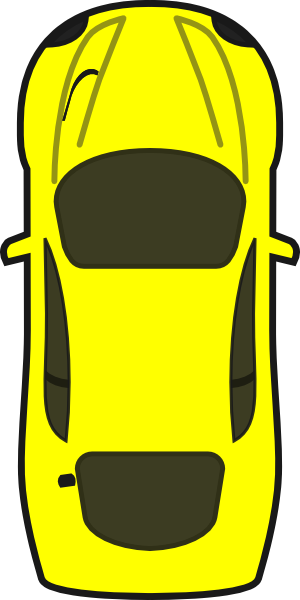} \vspace{2pt} \\
Human Vehicle & Autonomous Vehicle \vspace{2pt}
\end{tabular}
\centering
\includegraphics[width = 0.15\linewidth, angle = -90]{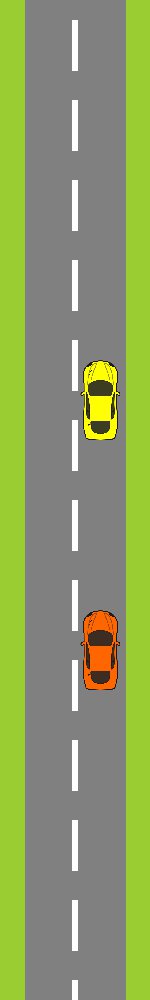}
\caption{Both vehicles currently travel to the right in the outer lane (lower). Autonomous vehicle (yellow) intends to influence human vehicle (orange) with intention to drive fast to inner lane (upper).}
\label{fig:intro}
\end{figure}

It is imperative for robots to behave reactively in a human-present environment because all safety specifications ought to be met. An autonomous vehicle, for instance, should yield to a human vehicle trying to nudge in front of it~\cite{lyu2021probabilistic,van2022provable}; a reconnaissance drone should avoid adversarial behaviors. Robots, however, are usually not designed to behave purely in a reactive manner because it makes them too conservative. Consider a scenario of autonomous driving (Fig.~\ref{fig:intro}) where the human vehicle is traveling in the outer lane (lower), but at a fast enough speed that it's more efficient to switch to the inner lane (upper). Many human drivers don't have the awareness to switch lanes because they are usually egocentric, even subconsciously, in that they would rather remain in their current lane unless blocked by some other vehicles. Such human egocentricity and strict infrastructure preconditions render purely communication-based approaches, like vehicle signaling or V2X~\cite{kato2002vehicle,baber2005cooperative,stubing2010simtd,de2014network,7355568}, fruitless in addressing these inefficiencies. Some works, therefore, proposed interaction-based approaches, like game-theoretical influence~\cite{toghi2021cooperative, sadigh2018planning}, wave stabilization~\cite{wu2018stabilizing, kreidieh2018dissipating,wu2017flow,kheterpal2018flow}, and herding~\cite{grover2022noncooperative, mohanty2023distributed, s17122729, pierson2017controlling, vaughan1998robot, vaughan2000experiments}, that use autonomous agents to influence human agents physically. In Fig.~\ref{fig:intro}, for instance, the autonomous vehicle would block the fast human vehicle, influencing it to drive in the inner lane.

Since such influence is exerted in close proximity, the autonomous agent needs an accurate human model. Although generally reasonable, models produced from offline techniques may not capture characteristics specific to the human agent with whom the robot interacts closely. 
For instance, in Fig.~\ref{fig:intro}, the autonomous vehicle is interested in, rather precisely, the human vehicle's desired travel velocity, in which each human differs from another.

Existing online approaches tackle this problem by positioning the autonomous agent as a passive observer, in which it observes the environmental states and their associated human response and then chooses the model that best explains this correlation. The issue with this design is that the autonomous agent is passive, so it only has access to information that the human agent chooses to exhibit. Hence, the autonomous agent can only make decisions based on the human information that’s readily available. In Fig.~\ref{fig:intro} for instance, a passive autonomous vehicle would presume the human vehicle intends to travel at most as fast as itself, whereas in reality the human could want to drive faster, only to be blocked by the autonomous vehicle.

In this work, we enable autonomous agents to leverage their actions to estimate the human internal model by actively interacting with the human to reveal more information. Rather than relying on passive observations, the autonomous agent can account for the fact that the human will react to its actions, so the autonomous agent can "probe", i.e., select actions that will trigger human reactions that will best challenge its initial belief. By probing iteratively, the autonomous agent converges to an increasingly accurate human model. Then, based on the probed information, the autonomous agent can actively influence other agents for some designated objectives, like higher efficiency or better driving experience. We propose our approach under some very mild assumptions, making it transferable to various human-robot interaction scenarios. Our key contributions include: 1) a coherent theoretical framework that unifies the probing and influencing procedures; 2) a proven solvable trajectory-planning optimization; 3) two case studies as application examples in the domain of autonomous driving with numerical simulations used to demonstrate the precision of probing results and efficacy in creating better participant driving experience during influence.

\section{Related Work}
To exert influence on humans, an autonomous agent would have to interact with different human agents in close proximity, who are heterogeneous agents that differ significantly in their internal models, to which the autonomous agent does not have direct access~\cite{stassen1990internal,jagacinski1978describing,lyu2022responsibility}. Such an internal model might characterize human's intentions, preferences, objectives, strategies, etc. Works in robotics and perception have focused on estimating these internal models using algorithms based on observations of human's actions, such as intent-driven behavior prediction~\cite{ziebart2009planning,liebner2012driver, dragan2012formalizing, awais2010human, baker2009action, nguyen2011capir, luce2012individual, lyu2020fg}, inverse reinforcement learning (IRL)~\cite{abbeel2005exploration,levine2012continuous, ng2000algorithms, ziebart2008maximum, ramachandran2007bayesian}, hidden model prediction~\cite{lam2015efficient, lyu2022adaptive}, affective state estimation~\cite{kulic2007affective}, and
activity recognition~\cite{van2008accurate}. Although the human model derived from the above methods performs generally reasonably, it might not capture specific characteristics of the human agent that the autonomous agent is interacting with. The autonomous agent, therefore, needs an online procedure to learn the model specially tailored to the human agent that the autonomous agent is interacting with.

Some online approaches frame this problem as a Partially Observable Markov Decision Process (POMDP)~\cite{javdani2015shared, fern2014decision,bandyopadhyay2013intention}, in which the autonomous agent parameterizes the human’s intent through a model, inferred through Markovian or Bayesian estimation of the hidden parameters of the internal models from observations of the physical states of the world~\cite{fisac2018probabilistically,hu2022sharp,tian2022safety, bhattacharyya2020online}. In this paradigm, the autonomous agent is mainly a reactive agent in the environment to observe, which sacrifices the robot's agency to initiate action to actively reveal information about the human. 

Some existing works enable active probing for interactive motion planning by incorporating a heuristic active information gathering objective, e.g., information entropy, into the autonomous agent's trajectory optimization framework for human value function parameter estimation~\cite{hu2022active, sadigh2016information}. Building upon this work, we allow the autonomous agent to optimize the information radius, i.e., the cohesion between two beliefs, relative to its latest belief of the human model. Then, instead of having a fixed reference belief as in~\cite{sadigh2016information}, the autonomous agent aims to maximize the information radius relative to a dynamic reference, its current belief, at every time iteration.

\section{Theory}
\label{section:theory}

\subsection{Human-Robot Joint Dynamics}
For all notations below, we use subscripts to denote the time step and superscripts to capture the attributes' ownership (human or robot). In a human-robot joint system, we define the state vector as $s_{t} \in \mathbb{R}^n$, the robot's input vector as $u^{\mathcal{R}}_{t} \in \mathbb{U}^{\mathcal{R}}  \subseteq \mathbb{R}^{m^{\mathcal{R}}}$, confined to admissible control space $\mathbb{U}^{\mathcal{R}}$, the human's input vector as $u^{\mathcal{H}}_{t} \in \mathbb{U}^{\mathcal{H}} \subseteq \mathbb{R}^{m^{\mathcal{H}}}$, confined to admissible control space $\mathbb{U}^{\mathcal{H}}$, and finally the discrete-time control-affine dynamics of the joint system as
\begin{equation}
\label{eq:dynamics}
s_{t+1} = f(s_{t}) + M^{\mathcal{R}}(s_{t})u^{\mathcal{R}}_{t} + M^{\mathcal{H}}(s_{t}) u^{\mathcal{H}}_{t}
\end{equation}
where $f: \mathbb{R}^n \to \mathbb{R}^n$ captures the non-linear autonomous dynamics and $M^{\mathcal{R}}: \mathbb{R}^n \to \mathbb{R}^{n \times m^{\mathcal{R}}}$ and $M^{\mathcal{H}}: \mathbb{R}^n \to \mathbb{R}^{n \times m^{\mathcal{H}}}$ are state-dependent input transformation matrices for robot and human respectively~\cite{hu2022active}.

\subsection{Belief Update}
The autonomous agent possesses a belief of $\varphi$ that characterizes the human agent's utility function $r^{\mathcal{H}}_{\varphi} : \mathbb{R}^n \to \mathbb{R}$. For driving scenarios, a typical $\varphi$ could characterize the desired velocity of the human vehicle, and a typical $r^{\mathcal{H}}_{\varphi}$ would include features like safety and speed. We generalize the autonomous agent's belief by proposing a non-parametric representation which can approximate a wider range of distributions. At time $t$, belief of $\varphi$ is defined as $bel_{t}$ with finite domain space $\Phi$.

The autonomous agent updates this belief via a particle-filtering recursion~\cite{thrun2002probabilistic}
\begin{equation}
\label{eq:beliefupdate} 
bel_{t+1}(\varphi) \propto bel_{t}(\varphi) \cdot
p(u^{\mathcal{H}}_t|s_t,u^{\mathcal{R}}_t,\varphi), \; \forall \varphi \in \Phi
\end{equation}
where the conditional probability is obtained through a softmax operation based on the Boltzmann model of exponential likeliness of human actions with greater utility~\cite{ziebart2008maximum, luce2012individual}
\begin{equation}
\begin{aligned}
\label{eq:boltzmann}
p(u^{\mathcal{H}}_t|s_t,u^{\mathcal{R}}_t,\varphi) = \frac{e^{\scriptscriptstyle r^{\mathcal{H}}_{\varphi} \left(f(s_{t}) + M^{\mathcal{R}}(s_{t})u^{\mathcal{R}}_{t} + M^{\mathcal{H}}(s_{t}) u^{\mathcal{H}}_{t}\right)}}{\sum_{\tilde{u}^{\mathcal{H}}_{t} \in \mathbb{U}^{\mathcal{H}}} e^{\scriptscriptstyle r^{\mathcal{H}}_{\varphi} \left(f(s_{t}) + M^{\mathcal{R}}(s_{t})u^{\mathcal{R}}_{t} + M^{\mathcal{H}}(s_{t}) \tilde{u}^{\mathcal{H}}_{t}\right)}}
\end{aligned}
\end{equation}
in which $\mathbb{U}^{\mathcal{H}}$ is discretized for softmax normalization. The complete belief update algorithm is shown in algorithm~\ref{alg:beliefupdate}.

\begin{algorithm}
\caption{Belief Update}\label{alg:beliefupdate}
\hspace*{\algorithmicindent} \textbf{Input: $bel_t, s_t, u^{\mathcal{R}}_{t}, u^{\mathcal{H}}_{t}$}
\begin{algorithmic}[1]
        \STATE $\eta \gets 0$
        \FORALL{$\varphi \in \Phi$}
            \STATE $r \gets e^{r^{\mathcal{H}}_{\varphi} \left(f(s_{t}) + M^{\mathcal{R}}(s_{t})u^{\mathcal{R}}_{t} + M^{\mathcal{H}}(s_{t}) u^{\mathcal{H}}_{t}\right)}$ \COMMENT{Boltzmann}
            
            \STATE $\tilde{r} \gets \sum_{\tilde{u}^{\mathcal{H}}_{t} \in \mathbb{U}^{\mathcal{H}}} e^{r^{\mathcal{H}}_{\varphi}\left(f(s_{t}) + M^{\mathcal{R}}(s_{t})u^{\mathcal{R}}_{t} + M^{\mathcal{H}}(s_{t}) \tilde{u}^{\mathcal{H}}_{t}\right)}$
            
            \STATE $bel_{t+1}(\varphi) \gets bel_{t}(\varphi) \cdot \frac{r}{\tilde{r}}$ \COMMENT{belief update}
            \STATE $\eta \gets \eta + bel_{t+1}(\varphi)$
        \ENDFOR
        \FORALL{$\varphi \in \Phi$}
            \STATE $bel_{t+1}(\varphi) \gets \frac{bel_{t+1}(\varphi)}{\eta}$ \COMMENT{belief normalization}
        \ENDFOR
        \RETURN $bel^{t+1}$             
\end{algorithmic}
\end{algorithm}

\subsection{Probing}
The motivation behind probing is to allow the autonomous agent to actively interact with the human agent to reveal more information that was previously unavailable, meaning that the autonomous agent should choose actions that best challenge its current belief at every time step. Quantitatively, the autonomous agent chooses actions that maximize the information radius between its current belief and the projected belief if such actions are to be executed.

We use Jensen-Shannon divergence (JSD) as a measure of information radius to quantify the cohesion between two beliefs, $bel_{a}$ and $bel_{b}$~\cite{lin1991divergence,sibson1969information}
\begin{equation}
\label{eq:jsd}
\begin{aligned}
D_{\mathrm{JS}}[bel_{a},bel_{b}]=\frac{D_{\mathrm{KL}}\left[bel_{a}:\overline{bel}_{a,b}\right ]+D_{\mathrm{KL}}\left[bel_{b}:\overline{bel}_{a,b}\right]}{2}
\end{aligned}
\end{equation}
where $D_{\mathrm{KL}}$ is the Kullback–Leibler divergence (KLD)~\cite{kullback1997information, cover1999elements} and $\overline{bel}_{a,b}$ is the arithmetic mixture of $bel_{a}$ and $bel_{b}$
\begin{equation}
\label{eq:kld}
\begin{aligned}
D_{\mathrm{KL}}[bel_{a}:\overline{bel}_{a,b}]=\underset{\varphi \sim bel_{a}}{\mathbb{E}} \log{\left(\frac{2 \cdot bel_{a}(\varphi)}{bel_{a}(\varphi)+bel_{b}(\varphi)}\right)}
\end{aligned}
\end{equation}

At state $s_{t}$, the autonomous agent predicts how the human agent, characterized by $\varphi$,  will react to its action $u^{\mathcal{R}}_{t}$ using
\begin{equation}
Q(s_{t}, u^{\mathcal{R}}_{t}, \varphi) = \underset{\tilde{u}^{\mathcal{H}}_{t} \in \mathbb{U}^{\mathcal{R}}}{\arg \max} \; r^{\mathcal{H}}_\varphi (f(s_{t}) + M^{\mathcal{R}}(s_{t})u^{\mathcal{R}}_{t} + M^{\mathcal{H}}(s_{t}) \tilde{u}^{\mathcal{H}}_{t}) \label{eq:probing_human}
\end{equation}

We solve the probing problem using Model Predictive Control (MPC), in which the autonomous agent chooses a sequence of actions that optimizes the JSD between the current belief and the projected belief at finite horizon $T$ 
\begin{subequations}
\label{eq:probing}
\begin{align}
\max_{u^{\mathcal{R}}_{0:T-1}} \; & \underset{\varphi \sim bel_{0}}{\mathbb{E}} \sum_{t=0}^{T-1} D_{\mathrm{JS}}[bel_{0},bel_{t+1}]-D_{\mathrm{JS}}[bel_{0},bel_{t}] \label{eq:probing_max}\\
\textrm{s.t.} \quad
& s_{0} = s_{t}, bel_{0} = bel_{t} \label{eq:probing_init}\\
& u^{\mathcal{H}}_{t} =  Q(s_{t}, u^{\mathcal{R}}_{t}, \varphi) \label{eq:probing_human}\\
& s_{t+1} = f(s_{t}) + M^{\mathcal{R}}(s_{t})u^{\mathcal{R}}_{t} + M^{\mathcal{H}}(s_{t}) u^{\mathcal{H}}_{t} \label{eq:probing_dynamics}\\
& bel_{t+1}(\varphi) \propto bel_{t}(\varphi) \cdot p(u^{\mathcal{H}}_t|s_t,u^{\mathcal{R}}_t,\varphi) \label{eq:probing_belief}
\end{align}
\end{subequations}

To ensure solvability, we prove that $D_{\mathrm{JS}}$, which maps to $[0,\infty)$ in theory, is upper-bounded in optimization~\eqref{eq:probing}.

\begin{proof}

\textbf{Boundedness}:

We first make a slight assumption that $bel_{0}$ is bounded and has compact support, hence
\begin{equation}
\label{eq:init}
\sup_{\varphi \in \Phi} bel_{0}(\varphi) < \infty  \land 
\inf_{\varphi \in \Phi} bel_{0}(\varphi)>0
\end{equation}
which helps to substantiate the boundedness of KLD~\cite{nielsen2020generalization}. We will initialize the belief such that condition~\eqref{eq:init} is satisfied in section~\ref{sec:simulation}.

We hypothesize inductively that $\forall a \in \{0,\dots,T-1\}$, $\sup_{\varphi \in \Phi} bel_{a}(\varphi) < \infty$. Since $p(u^{\mathcal{H}}_t|s_t,u^{\mathcal{R}}_t,\varphi)$ maps to an image of $(0,1)$, using condition~\eqref{eq:init} as base case, we have
\begin{equation}
\label{eq:support}
\sup_{\varphi \in \Phi} bel_{a}(\varphi)<1<\infty, \; \forall a \in \{0,\dots,T\}
\end{equation}

By similar induction technique, we have
\begin{equation}
\label{eq:support}
\inf_{\varphi \in \Phi} bel_{a}(\varphi)>0, \; \forall a \in \{0,\dots,T\}
\end{equation}

Hence, we have extended condition~\eqref{eq:init} to
\begin{equation}
\label{eq:lowerupper}
\sup_{\varphi \in \Phi} bel_{a}(\varphi)<\infty\land
\inf_{\varphi \in \Phi} bel_{a}(\varphi)>0, \; \forall a \in \{0,\dots,T\}
\end{equation}

Therefore, $\forall a \in \{0,\dots,T\}$, $\exists \bar{s}=\sup_{\varphi \in \Phi} bel_{a}(\varphi)$ such that $0<\bar{s}<\infty$. Similarly, $\forall a,b \in \{0,\dots,T\}$, $\exists \underbar{i}=\inf_{\varphi \in \Phi} bel_{a}(\varphi) + bel_{b}(\varphi)$ such that $0< \underbar{i}<\infty$.

Therefore, by equation~\eqref{eq:kld}, we have $\forall a,b \in \{0,\dots,T\}$
\begin{equation}
\label{eq:abscont}
\begin{aligned}
D_{\mathrm{KL}}[bel_{a}:\overline{bel}_{a,b}]&=\underset{\varphi \sim bel_{a}}{\mathbb{E}} \log{\left(\frac{2 \cdot bel_{a}(\varphi)}{bel_{a}(\varphi)+bel_{b}(\varphi)}\right)}\\
& \leq  \underset{\varphi \sim bel_{a}}{\mathbb{E}} \sup_{\varphi \in \Phi} \log\left(\frac{2 \cdot bel_{a}(\varphi)}{bel_{a}(\varphi)+bel_{b}(\varphi)}\right)\\
& \leq \log(2 \cdot \bar{s}) - \log(\underbar{i})<\infty
\end{aligned}
\end{equation}

By symmetry, $D_{\mathrm{KL}}[bel_{b}:\overline{bel}_{a,b}]<\infty$ can be easily proved using the same technique, which together concludes the boundedness of JSD.
\end{proof}

We adopt a dynamic-programming-based approach to optimize equation~\eqref{eq:probing}, while other quasi-Newton methods, like the BFGS algorithm~\cite{broyden1970convergence,fletcher1970new,goldfarb1970family,shanno1970conditioning}, are also applicable. Although the computational complexity grows exponentially with respect to the state dimension, the high parallelizability of equation~\eqref{eq:probing_dynamics} and~\eqref{eq:probing_belief} can attenuate the curse of dimensionality. Moreover, we argue that successfully reasoning about human-robot interactions over a short horizon does not require a full-fidelity model of the joint dynamics, so highly informative insights can still be obtained tractably via approximation. We define the value function of executing $n$ consecutive controls starting from time $k$ as
\begin{equation}
\label{eq:bellmandefine}
V(k,n) = \underset{\varphi \sim bel_{0}}{\mathbb{E}} \sum_{t=k}^{k+n-1} D_{\mathrm{JS}}[bel_{0},bel_{t+1}]-D_{\mathrm{JS}}[bel_{0},bel_{t}]
\end{equation}
The value function on the horizon therefore satisfies
\begin{equation}
\label{eq:proof}
\begin{aligned}
V(0,T)
&=\underset{\varphi \sim bel_{0}}{\mathbb{E}}\sum_{t=0}^{k-1} D_{\mathrm{JS}}[bel_{0},bel_{t+1}]-D_{\mathrm{JS}}[bel_{0},bel_{t}] \\
&+ \underset{\varphi \sim bel_{0}}{\mathbb{E}} \sum_{t=k}^{T-1} D_{\mathrm{JS}}[bel_{0},bel_{t+1}]-D_{\mathrm{JS}}[bel_{0},bel_{t}]\\
& = V(0, k) + V(k, T-k), \; \forall k \in \{0,\dots,T\}
\end{aligned}
\end{equation}
which shows that the path-dependency fits a Bellman optimality equation~\cite{bellman2015applied}. The optimal value function and control policy can therefore be obtained in polynomial time by backtracking the Hamilton–Jacobi–Bellman (HJB) equation~\cite{bellman1966dynamic}
\begin{equation}
\label{eq:hjb}
\begin{aligned}
V(t, T-t) = \max_{u^{\mathcal{R}}_{t} \in \mathbb{U}^{\mathcal{R}}}\biggl\{V(t,1)+V(t+1, T-t-1)\biggl\}\\
\end{aligned}
\end{equation}

Following this policy, the autonomous agent interactively probes the human agent and gradually converges its belief until the change of JSD is too small. The autonomous agent then chooses $\hat{\varphi}$, which could be a linear combination of all $\varphi \in \Phi$ weighted by their $bel(\varphi)$ or simply the most likely $\varphi \in \Phi$, as the human model parameter.

\subsection{Influence}
We characterize an influence as a sequence of atomic objectives, each with a utility function, that accounts for a major influence if all executed in order, and we delegate the responsibility of planning these atomic objectives to some high-level planner. For each objective, we incorporate $\hat{\varphi}$ into the utility function for both robot and human.
\begin{subequations}
\label{eq:influence}
\begin{align}
\max_{u^{\mathcal{R}}_{0:T-1}} \; &  \sum_{t=0}^{T-1} r^{\mathcal{R}}_{\hat{ \varphi}}(s_{t+1}) \label{eq:influence_max}\\
\textrm{s.t.} \quad
& s_{0} = s_{t} \label{eq:influence_init}\\
& u^{\mathcal{H}}_{t} =  Q(s_{t}, u^{\mathcal{R}}_{t}, \hat{\varphi}) \label{eq:influence_human}\\
& s_{t+1} = f(s_{t}) + M^{\mathcal{R}}(s_{t})u^{\mathcal{R}}_{t} + M^{\mathcal{H}}(s_{t}) u^{\mathcal{H}}_{t} \label{eq:influence_dynamics}
\end{align}
\end{subequations}

Similarly, this optimization problem can be solved using HJB recursion in polynomial time.

\section{Simulation}
\label{sec:simulation}
In this section, we present two car-following-based scenarios in which probing and influencing can be used to facilitate better participant experience and optimality for human drivers. Both scenarios start with the human vehicle following the autonomous vehicle.

\subsection{Ground Truth}
To generate the ground truth trajectories for the human-driven vehicle, we use the intelligent driver model (IDM)~\cite{treiber1999explanation, treiber2013traffic, treiber2017intelligent}, which is known to accurately imitate human driving behaviors.
\begin{equation}
\begin{aligned}
\label{eq:idm}
u^{\mathcal{H}}=u_{\mathrm{max}}\left[1- \left (\frac{v^\mathcal{H}}{v_\mathrm{des}} \right)^4 - \left (\frac{d_\mathrm{des}}{ x^{\mathcal{R}}-x^{\mathcal{H}} } \right)^2 \right]
\end{aligned}
\end{equation}
in which
\begin{equation}
\begin{aligned}
\label{eq:idm}
d_\mathrm{des}= d_\mathrm{min} + \tau_{\mathrm{gap}} \cdot v^\mathcal{H} - \frac{v^\mathcal{H} \cdot ( v^{\mathcal{H}}  -  v^{\mathcal{R}}  )}{2 \sqrt{a_\mathrm{max} \cdot b_\mathrm{pref}}}
\end{aligned}
\end{equation}
where superscripted notations are system dynamics and subscripted notations are constant parameters. We assume that humans will maintain their driving style, so the constant parameters are static over time. Without loss of generality, we also use IDM to model other background vehicles in the environment. We simulate all vehicles as mass points.

\subsection{Exploitation and Exploration}
To balance exploitation and exploration, the autonomous vehicle alternates between \SI{5}{\second} of passive observation and \SI{5}{\second} of active probing. We also set the MPC horizon to \SI{5}{\second}. Thanks to the boundedness of JSD, we can add a safety objective, $\lambda \cdot r^{\mathcal{R}}_{safe}(s_{t+1})$, on the autonomous agent's optimization to enforce some safety features, and we choose $\lambda$ empirically.

\subsection{Human Model}
The autonomous vehicle models the human underlying utility using a combination of features, namely desired headway and desired velocity. For each scenario, we choose $\lvert \Phi \rvert=30$ such that each $\varphi \in \Phi$ maps to a distinct desired velocity or desired headway, and we initialize them to a uniform distribution, which satisfies condition~\eqref{eq:init}.

\subsection{Scenario 1: Influence fast drivers to switch lane}
\begin{figure}
    \centering
    \begin{subfigure}{0.15\linewidth}
        \centering
        \includegraphics[width = \linewidth]{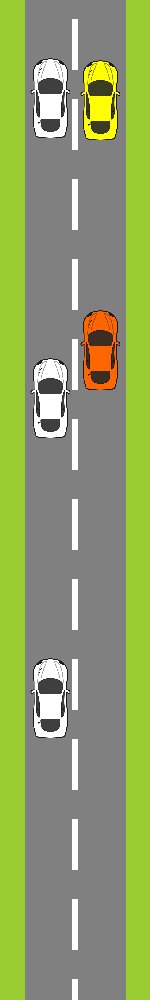}
        \caption{Phase 1}
        \label{fig:scenario1phase1}
    \end{subfigure}
    \hspace{1em}
    \begin{subfigure}{0.15\linewidth}
        \centering
        \includegraphics[width = \linewidth]{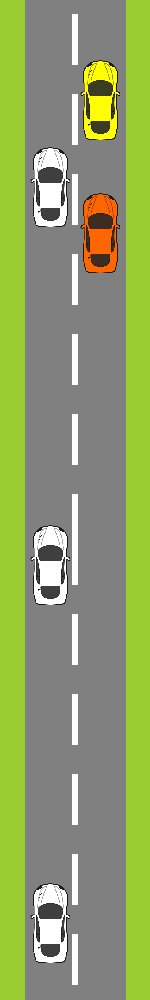}
        \caption{Phase 2}
        \label{fig:scenario1phase2}
    \end{subfigure}
    \hspace{1em}
    \begin{subfigure}{0.15\linewidth}
        \centering
        \includegraphics[width = \linewidth]{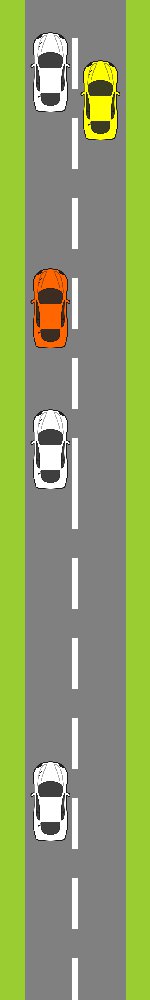}
        \caption{Phase 3}
        \label{fig:scenario1phase3}
    \end{subfigure}
    \begin{subfigure}{0.4\linewidth}
        \centering
        \includegraphics[width = \linewidth]{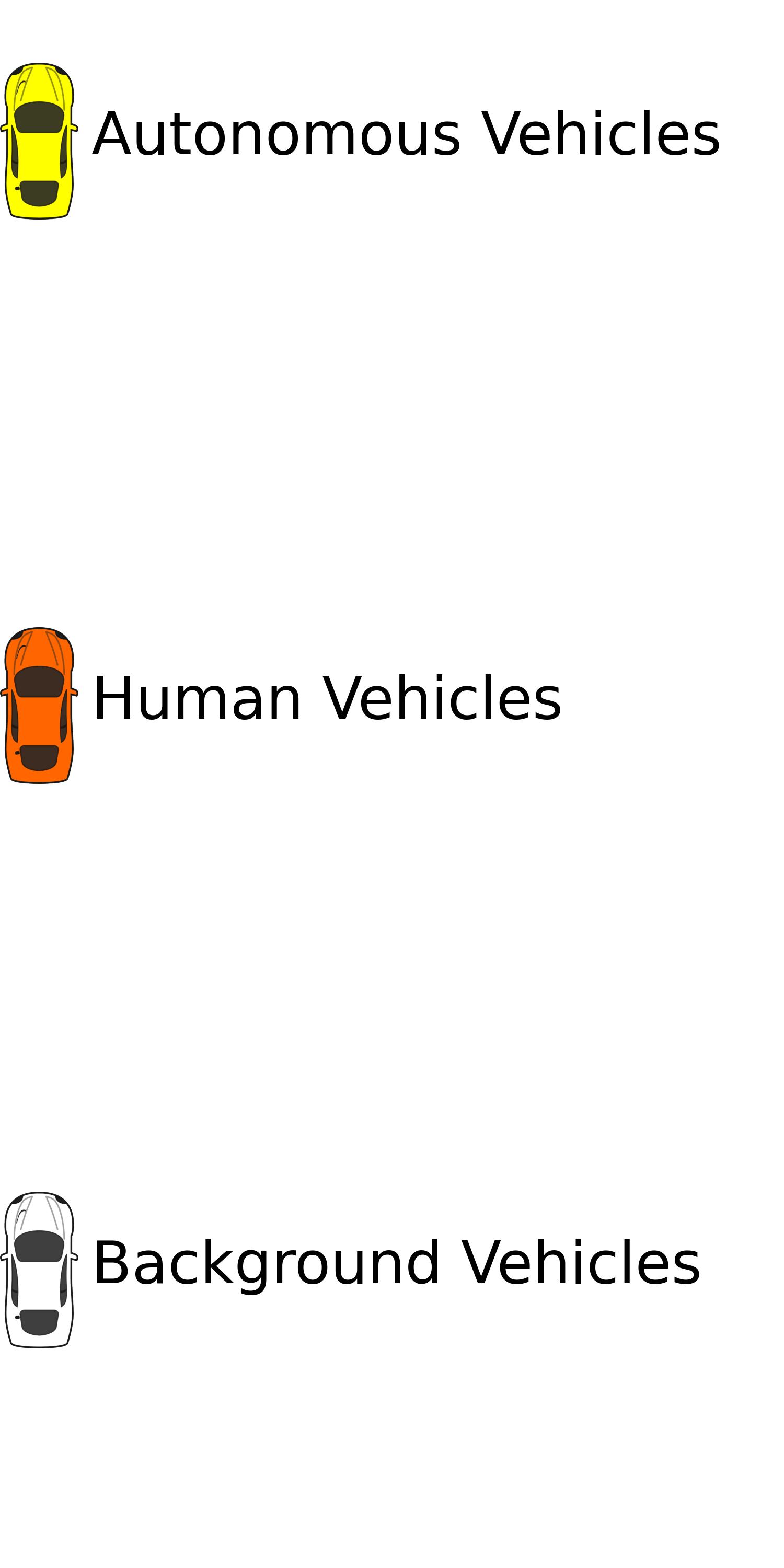}
        \label{fig:scenario1legend}
    \end{subfigure}
    \vspace{15pt}
  \caption{Phase 1: Autonomous vehicle maintains velocity. Phase 2: Autonomous vehicle brakes to block the human vehicle. Phase 3: Human vehicle merges due to blocking. All vehicles travel upwards.}
  \label{fig:scenario1}
\end{figure}
Consider a two-lane highway (Fig.~\ref{fig:scenario1phase1}) with an inner lane (left) and an outer lane (right).
Here, we cause the autonomous vehicle to actively probe the desired velocity of the human vehicle. If the human vehicle exhibits the intention to travel at a high velocity, the autonomous vehicle will perform a series of maneuvers to help the human vehicle merge to the inner lane in the widest gap between the background vehicles. While approaching the widest gap, the autonomous vehicle slows down to block the human vehicle (Fig.~\ref{fig:scenario1phase2}), and the human vehicle switches lanes shortly after that (Fig.~\ref{fig:scenario1phase3}).

\begin{figure}
\centering
\includegraphics[width=.5\textwidth]{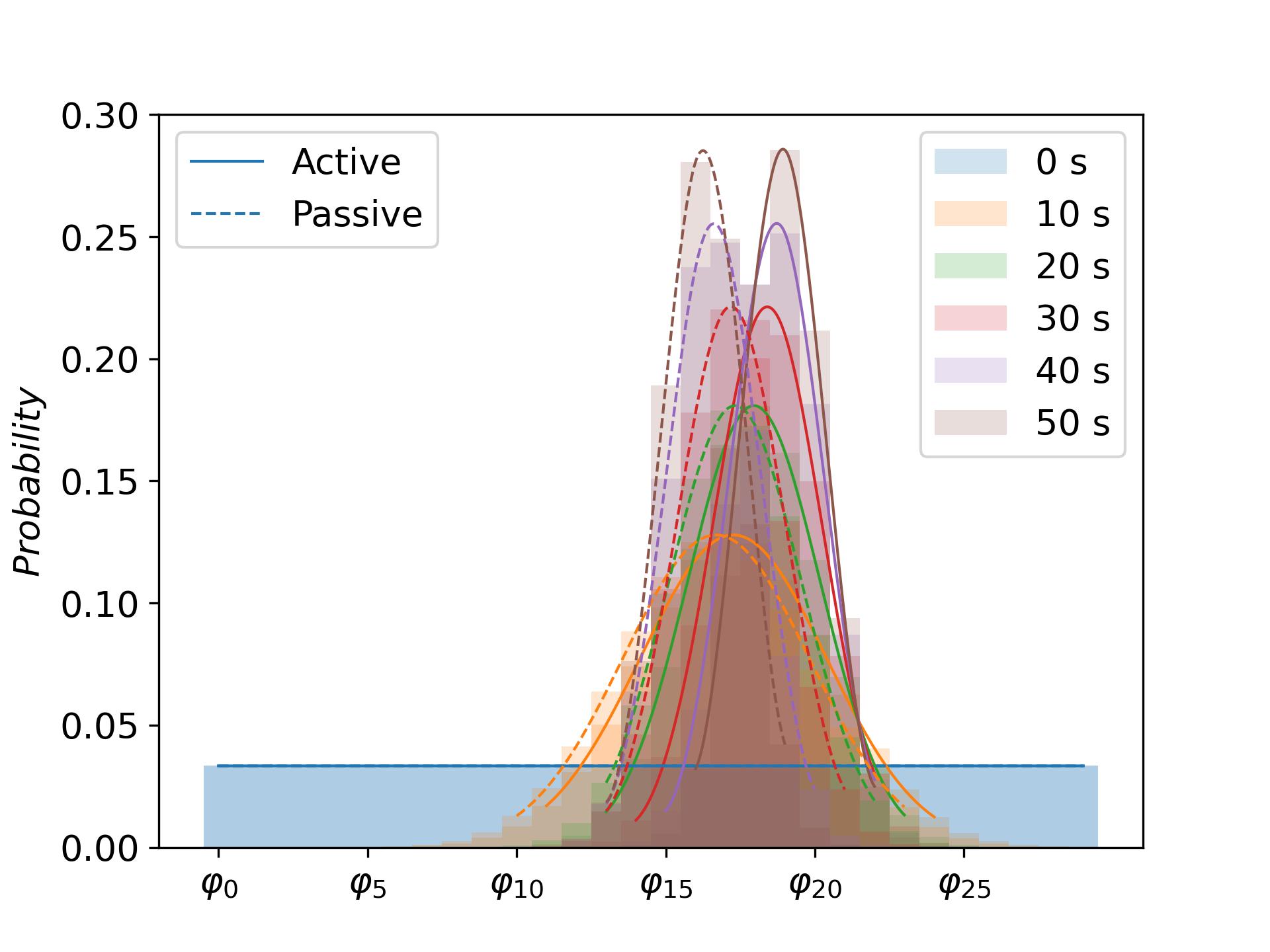}
\caption{Belief Snapshot}
\label{fig:snapshot_1}
\end{figure}

We choose the IDM parameters as $u_\mathrm{max}=\SI{0.73}{\meter/ \second^2}$, $b_\mathrm{pref}=\SI{1.67}{\meter/ \second^2}$, $v_\mathrm{des}=\SI{25}{\meter/ \second}$, $\tau_{\mathrm{gap}}=\SI{1.5}{\second}$, and $d_\mathrm{min}=\SI{2}{\meter}$. We start the car-following scenario with relative headway of \SI{100}{\meter}, the autonomous vehicle and the human vehicle both traveling at \SI{20}{\meter/ \second}. We also included a passive observing approach to compare as a baseline. Fig.~\ref{fig:snapshot_1} is a snapshot of the belief from two approaches taken every \SI{10}{\second}.

By \SI{50}{\second}, the active approach's peak happens at $\varphi_{19}$, which maps to a desired velocity of \SI{23.56}{\meter/\second}, which is close to the IDM parameter, $v_\text{des}$, of \SI{25}{\meter/\second}. In comparison, the passive observation baseline peaks at $\varphi_{16}$ that maps to \SI{19.86}{\meter/\second}, which is very far from the ground truth. This is because the passive approach suffers from no exploration to trigger human reaction, so the autonomous vehicle will assume the human vehicle intends to travel only as fast as itself.

\begin{figure}
\centering
\includegraphics[width=.5\textwidth]{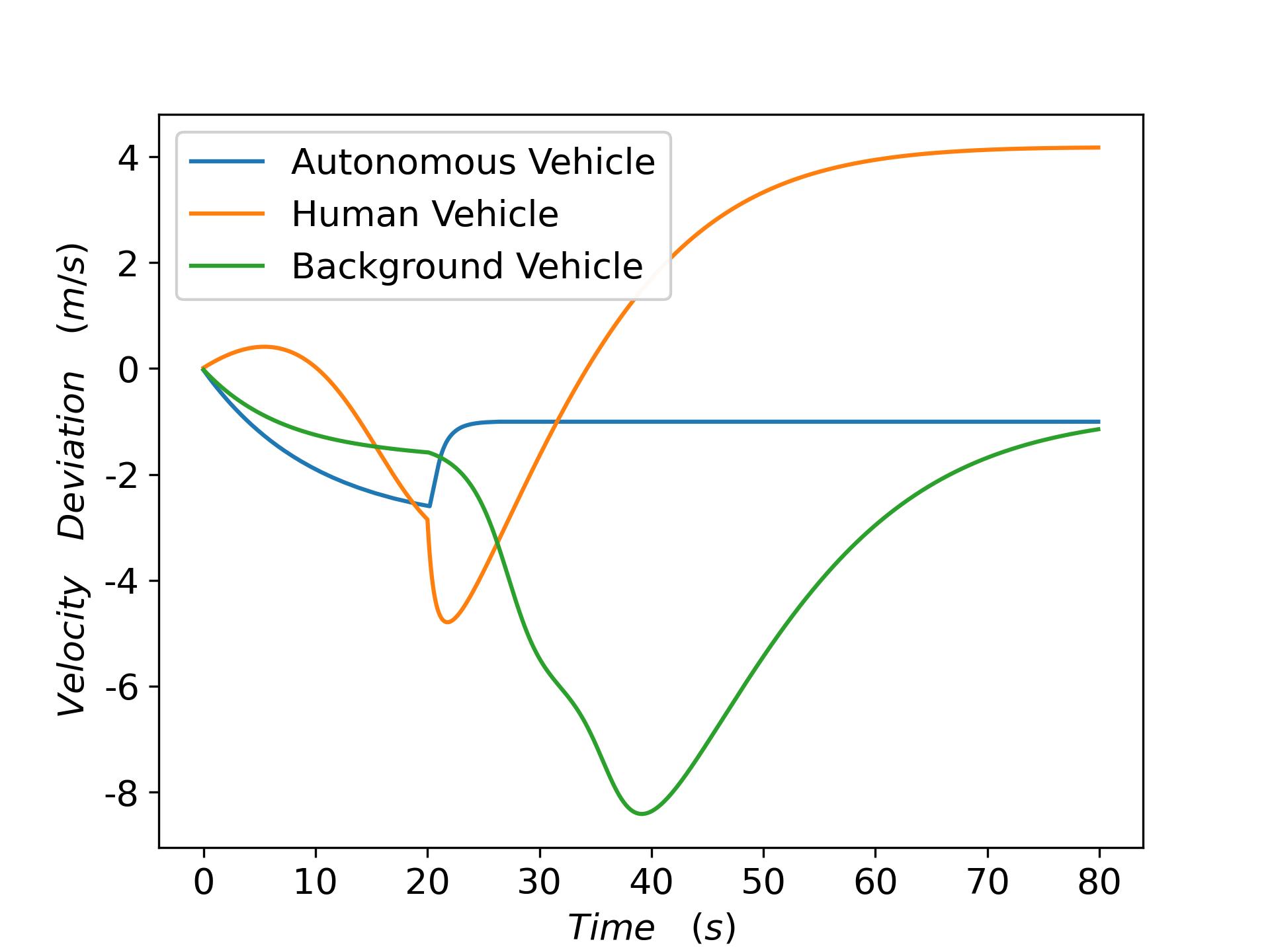}
\caption{Velocity Deviation}
\label{fig:velocity_1}
\end{figure}

\begin{figure}
\centering
\includegraphics[width=.5\textwidth]{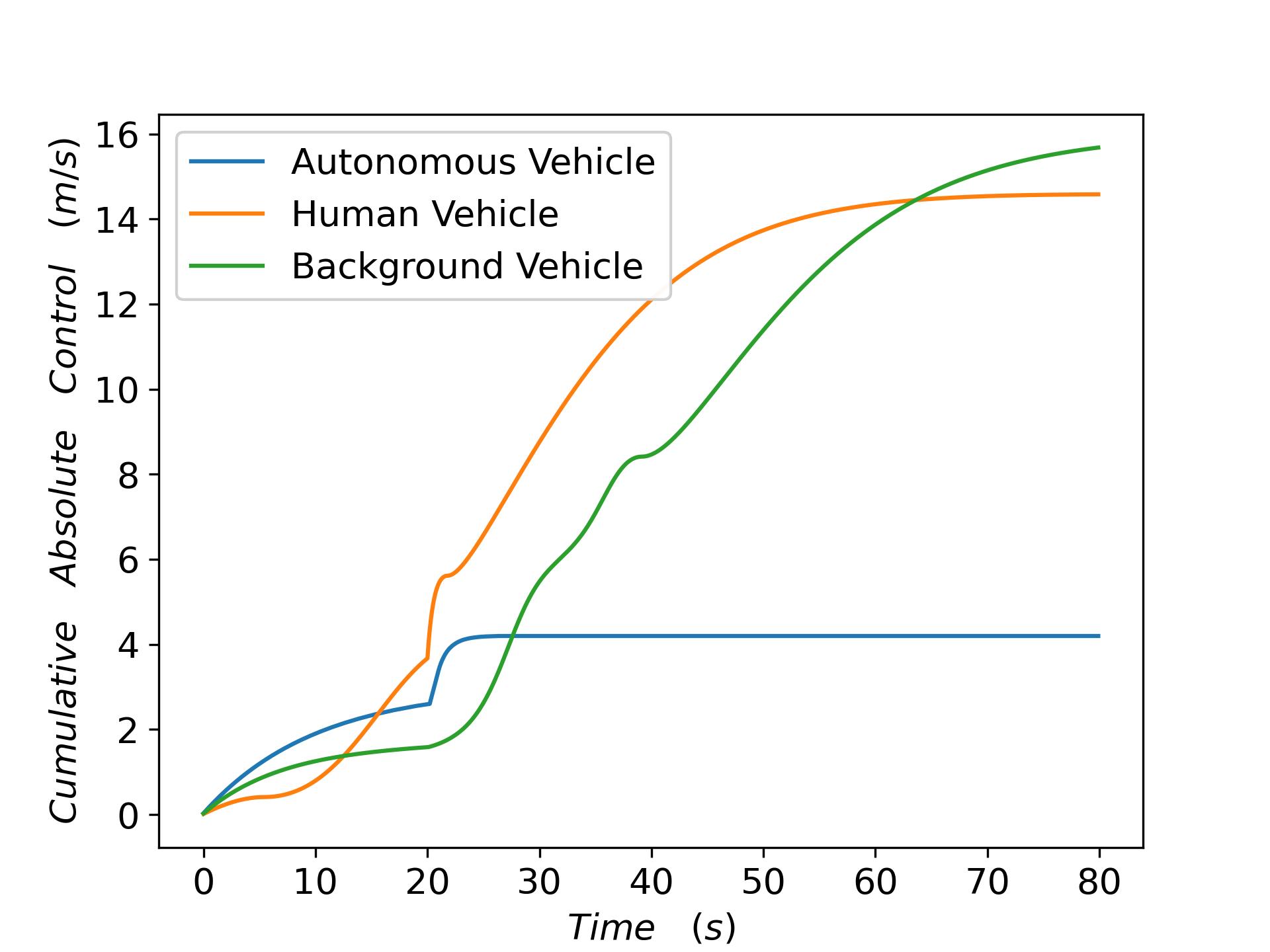}
\caption{Cumulative Absolute Control}
\label{fig:deviation_1}
\end{figure}

Leveraging this probed information, the autonomous vehicle can set a cutoff, \SI{23}{\meter/\second} in our simulation for instance, to influence the humans with high desired velocity to drive in the inner lane. According to Fig.~\ref{fig:velocity_1}, the influence brought about a 20.04\% increase in the human vehicle's velocity, whereas the passive approach wouldn't be able to initiate the influence procedure at all because it does not try to elicit information that the human is not providing, hence the autonomous vehicle becomes more and more wrongly convinced that the human vehicle intends to travel only as fast as \SI{19.86}{\meter/\second}. According to Fig.~\ref{fig:deviation_1}, the influence introduces a bounded perturbation, about~\SI{15.68}{\meter/\second} of cumulative absolute control, on average background vehicles, which could be easily attenuated with autonomous vehicles using flow stopper techniques~\cite{sugiyama2008traffic, wu2018stabilizing}.

\subsection{Scenario 2: Helping human to switch lane}
\begin{figure}
    \centering
    \begin{subfigure}{0.15\linewidth}
        \centering
        \includegraphics[width = \linewidth]{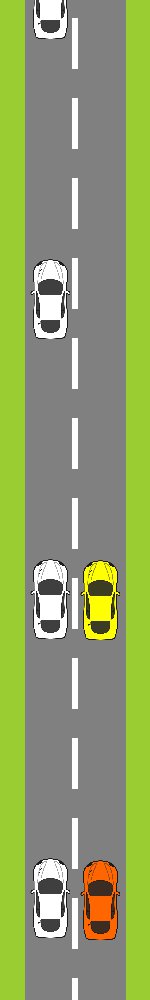}
        \caption{Phase 1}
        \label{fig:scenario2phase1}
    \end{subfigure}
    \hspace{1em}
    \begin{subfigure}{0.15\linewidth}
        \centering
        \includegraphics[width = \linewidth]{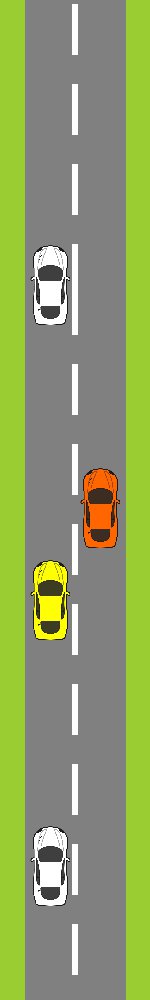}
        \caption{Phase 2}
        \label{fig:scenario2phase2}
    \end{subfigure}
    \hspace{1em}
    \begin{subfigure}{0.15\linewidth}
        \centering
        \includegraphics[width = \linewidth]{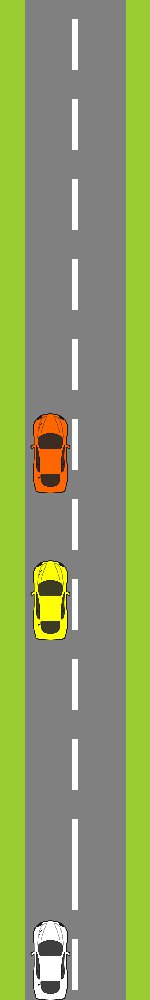}
        \caption{Phase 3}
        \label{fig:scenario2phase3}
    \end{subfigure}
    \begin{subfigure}{0.4\linewidth}
        \centering
        \includegraphics[width = \linewidth]{legend.jpg}
        \label{fig:scenario2legend}
    \end{subfigure}
    \vspace{15pt}
  \caption{Phase 1: Autonomous vehicle merges first. Phase 2: Autonomous vehicle slows down to create gap for human vehicle. Phase 3: Human vehicle merges. All vehicles travel upwards.}
  \label{fig:scenario2}
\end{figure}

Consider a scenario like Fig.~\ref{fig:scenario2phase1}, in which the lane the autonomous and human vehicle currently occupy is about to end, either due to traffic, construction, or lane merge. Both vehicles, therefore, have to switch to the left lane, which is occupied by some background vehicles. Assume the headway gaps between the background vehicles are too narrow for humans while traveling at such a high speed. Fortunately, autonomous vehicles are capable of performing the switching. The autonomous vehicle, therefore, helps the human vehicle to switch lanes by first probing the desired headway of the human vehicle around a specific velocity, in this case \SI{20}{\meter/\second}. The autonomous vehicle will then switch lanes and slow down to create enough gap based on the probed headway (Fig.~\ref{fig:scenario2phase2}). Finally, the human vehicle can merge into the lane with ease (Fig.~\ref{fig:scenario2phase3}).

We choose the IDM parameters as $u_\mathrm{max}=\SI{0.73}{\meter/ \second^2}$, $b_\mathrm{pref}=\SI{1.67}{\meter/ \second^2}$, $v_\mathrm{des}=\SI{20}{\meter/ \second}$, $\tau_{\mathrm{gap}}=\SI{1.5}{\second}$, and $d_\mathrm{min}=\SI{2}{\meter}$. Similarly, we initialize the road condition to the same condition as the previous scenario, and we include a passive observing approach to compare as a baseline.

\begin{figure}
\centering
\includegraphics[width=.5\textwidth]{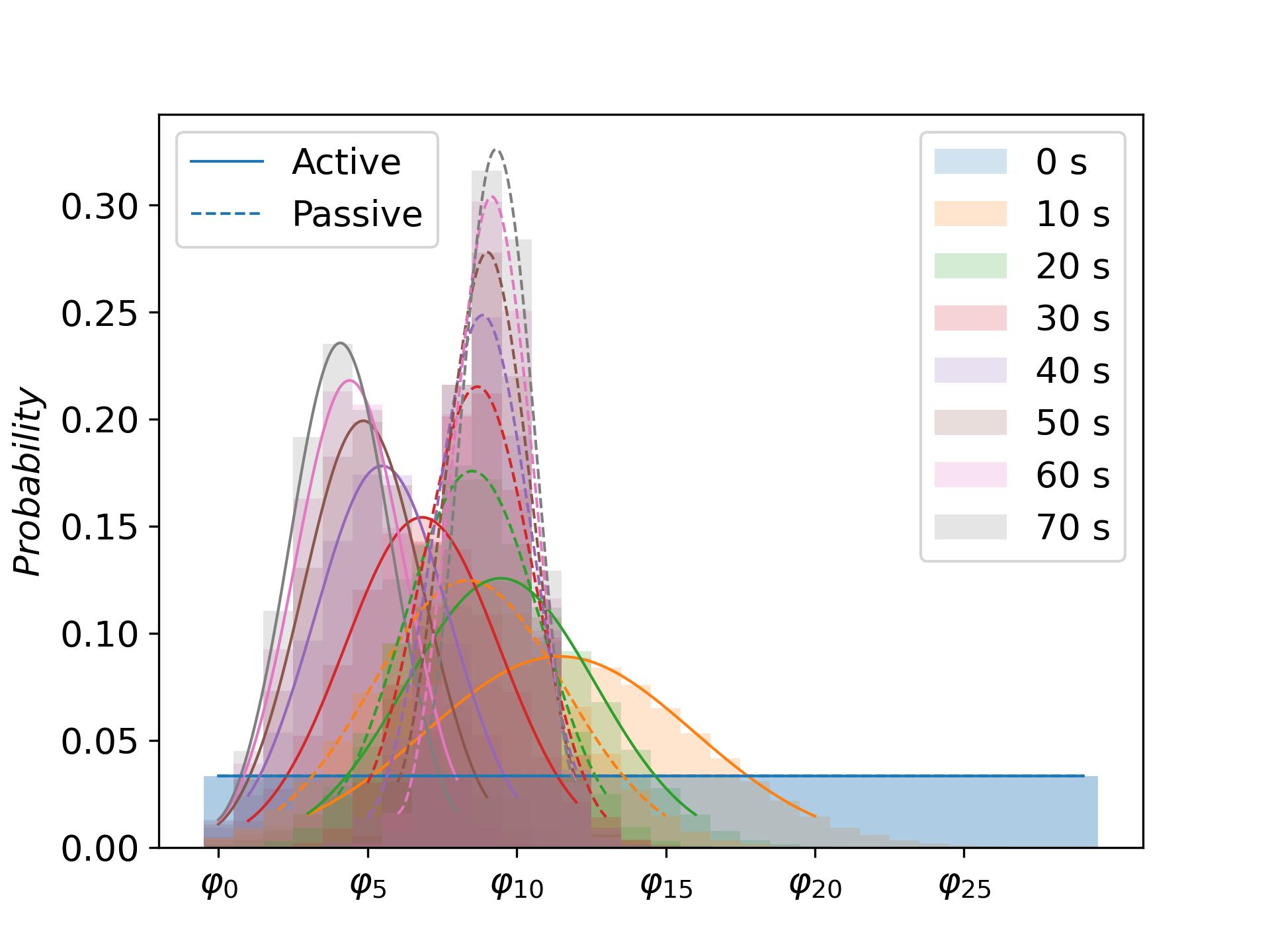}
\caption{Belief Snapshot}
\label{fig:snapshot_2}
\end{figure}

Fig.~\ref{fig:snapshot_2} is a snapshot of the belief from two approaches taken every \SI{10}{\second}. By \SI{70}{\second}, the probability for the active approach peaks at $\varphi_{4}$, which maps to a desirable headway around \SI{48.27}{\meter}, whereas that of passive approach peaks at $\varphi_{9}$, which maps to a desirable headway around \SI{108.62}{\meter}. For reference, according to data from the Next Generation Simulation for US Highway 101~\cite{colyar2007us}, the average headway for cars traveling around \SI{20}{\meter/ \second} is about \SI{42.18}{\meter}. Although not absolutely precise, the active approach generates a much more accurate profile than the passive approach does.

\begin{figure}
\centering
\includegraphics[width=.5\textwidth]{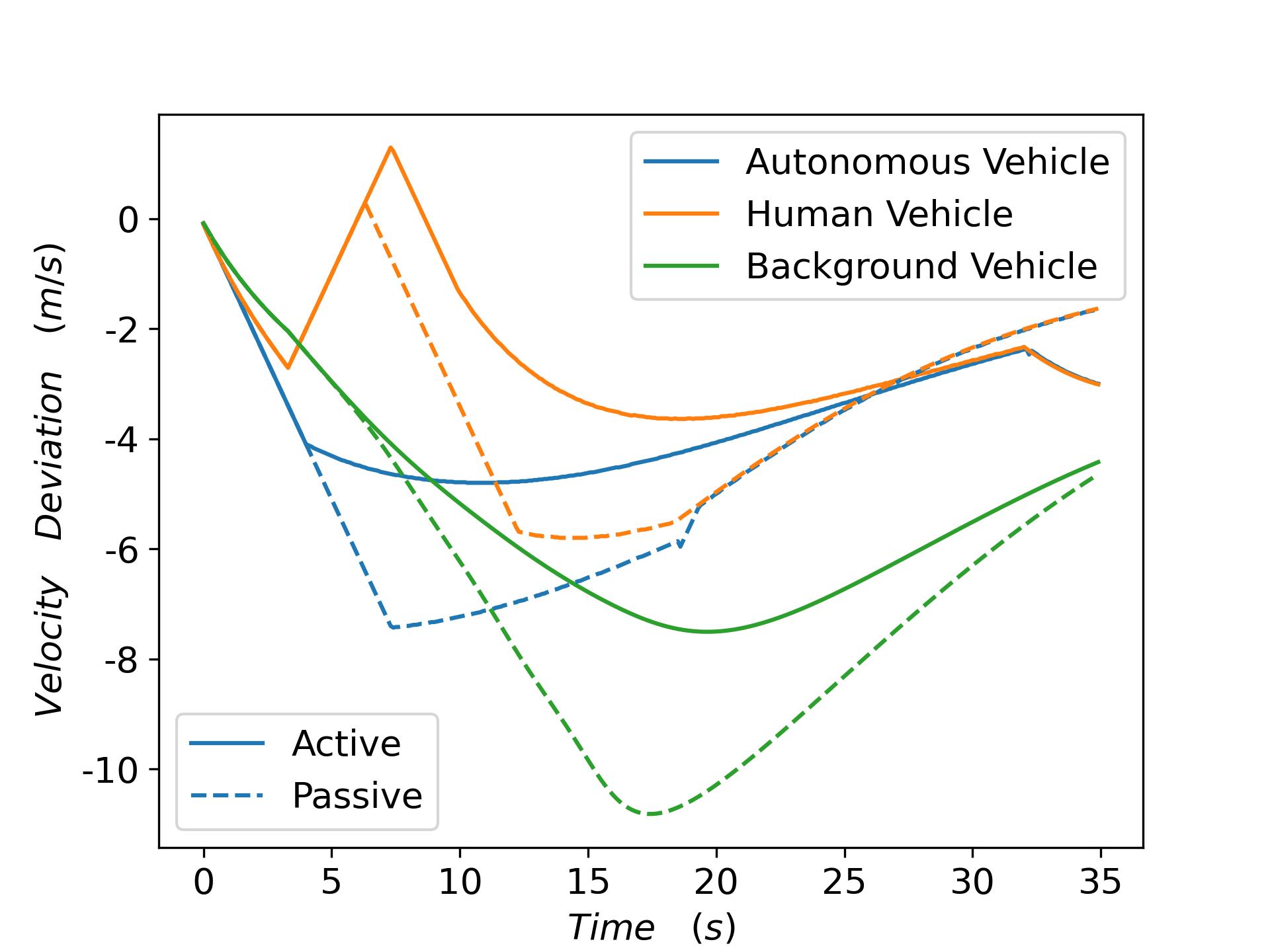}
\caption{Velocity Deviation}
\label{fig:velocity_2}
\end{figure}

\begin{figure}
\centering
\includegraphics[width=.5\textwidth]{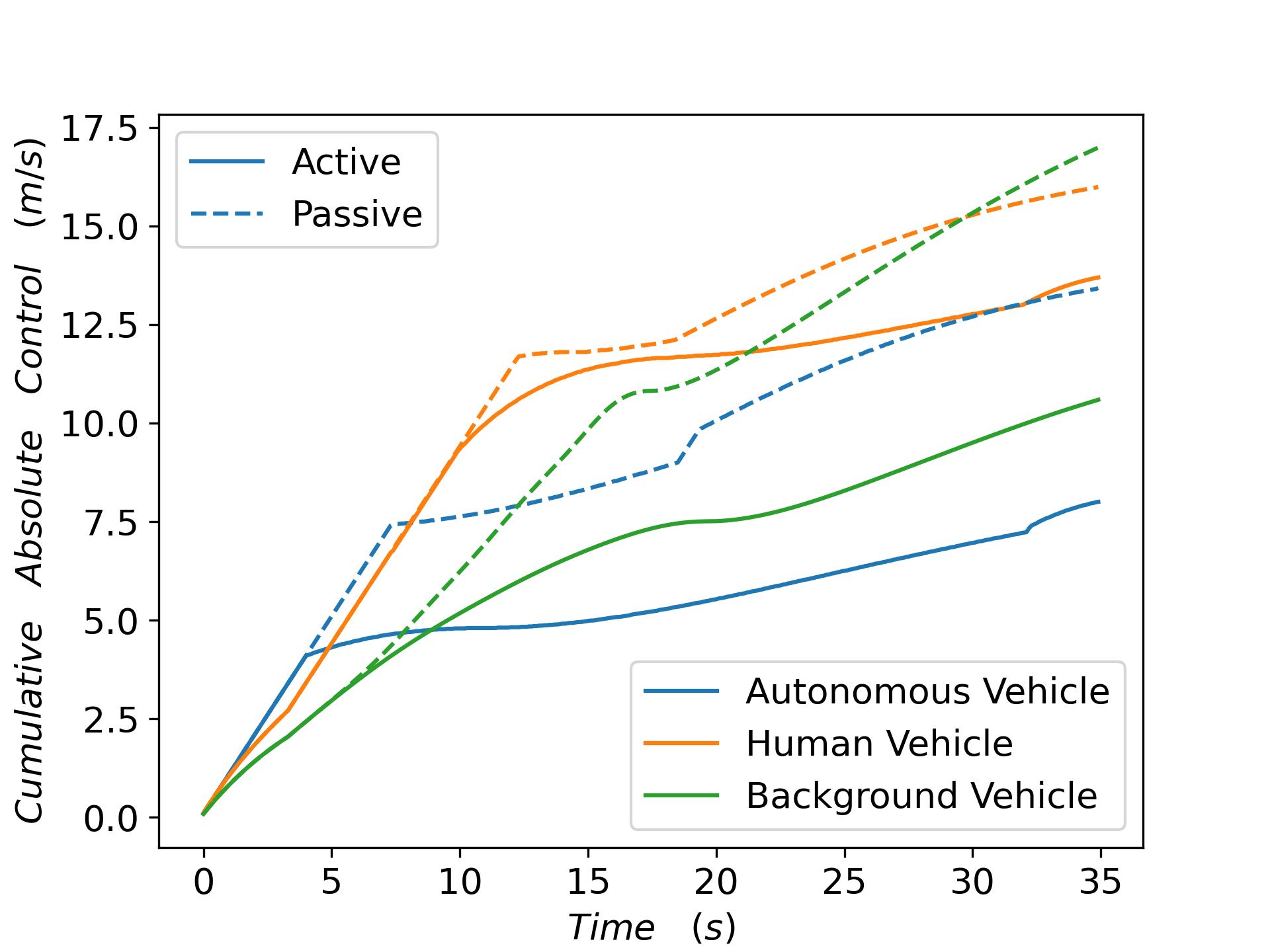}
\caption{Cumulative Absolute Control}
\label{fig:deviation_2}
\end{figure}

Based on the probed information, the autonomous vehicle can proceed to create a gap for the human vehicle. For comparison, we simulated a baseline where the autonomous vehicle is passive during the information gathering process, so the autonomous vehicles would have to slow down to create a wider gap, inducing larger perturbations on the background vehicles. According to Fig.~\ref{fig:deviation_2}, the cumulative absolute control for all three types of vehicles in the active approach is significantly lower than that in the passive approach. The reductions in perturbation are respectively $40.36\%$, $14.33\%$, and $37.66\%$ for autonomous, human, and background vehicle. According to Fig.~\ref{fig:velocity_2}, the active approach generates less extreme velocity deviation for all three types of vehicles in general, which helps to reduce the intensity and propagation of traffic wave~\cite{8569485}.

Moreover, our baseline is under the assumption that the autonomous vehicle would overtake under this scenario. Without active probing, the autonomous vehicle is more likely to behave quite conservatively, so it will most likely wait until all of the background vehicles have passed to switch lanes. This subjects the autonomous and human vehicles to almost a complete stop and a wait time that depends on the number of consecutive closely spaced background vehicles behind, meaning that the deviation will continue to increase if there is no large gap. Our active probing and influencing approach, on the other hand, is agnostic to this condition because the autonomous vehicle creates its own lane-change opportunity.

\section{Conclusions}
In this work, we present an active probing approach for an autonomous
agent to actively interact with a human agent to reveal information about a human's underlying utility and internal model. Our simulation results in autonomous driving demonstrate how the gathered information can be leveraged to increase driver experience and overall optimality compared to a passive learning baseline method. Future work could adopt learning-based methods to replace the heuristic probing objective with a more efficient and scenario-specific objective. It could also be worthwhile to relax the static human model assumption, hence empowering the autonomous agent to actively learn the human's adaptation policy.

\section{Acknowledgement}
This work was supported by the CMU Argo AI Center for Autonomous Vehicle Research.

Special thanks to Mrinal Verghese and Bhaskar Krishnamachari for their insightful suggestions, Rachel Burcin for her hospitality, and the 2022 Robotics Institute Summer Scholars for their company.

\newpage
\balance
\bibliographystyle{IEEEtran}
\bibliography{main}

\end{document}